\documentclass[10pt,journal,compsoc]{IEEEtran}

\usepackage{caption}
\usepackage{subcaption}
\usepackage{graphicx}
\usepackage{pgfplots}
\usepackage{booktabs}
\usepackage{multirow}
\usepackage{amsmath, amssymb}

\ifCLASSOPTIONcompsoc
  \usepackage[nocompress]{cite}
\else
  \usepackage{cite}
\fi
\ifCLASSINFOpdf
\else
\fi
\begin{document}
%
% paper title
% Titles are generally capitalized except for words such as a, an, and, as,
% at, but, by, for, in, nor, of, on, or, the, to and up, which are usually
% not capitalized unless they are the first or last word of the title.
% Linebreaks \\ can be used within to get better formatting as desired.
% Do not put math or special symbols in the title.
%\title{A Survey on How to Leverage multimodal Learning to Wearable Sensor-based Human Action Recognition}

\title{KEDformer:Knowledge Extraction Seasonal Trend Decomposition for Long-term Sequence Prediction}

%
%
% author names and IEEE memberships
% note positions of commas and nonbreaking spaces ( ~ ) LaTeX will not break
% a structure at a ~ so this keeps an author's name from being broken across
% two lines.
% use \thanks{} to gain access to the first footnote area
% a separate \thanks must be used for each paragraph as LaTeX2e's \thanks
% was not built to handle multiple paragraphs
%
%
%\IEEEcompsocitemizethanks is a special \thanks that produces the bulleted
% lists the Computer Society journals use for "first footnote" author
% affiliations. Use \IEEEcompsocthanksitem which works much like \item
% for each affiliation group. When not in compsoc mode,
% \IEEEcompsocitemizethanks becomes like \thanks and
% \IEEEcompsocthanksitem becomes a line break with idention. This
% facilitates dual compilation, although admittedly the differences in the
% desired content of \author between the different types of papers makes a
% one-size-fits-all approach a daunting prospect. For instance, compsoc 
% journal papers have the author affiliations above the "Manuscript
% received ..."  text while in non-compsoc journals this is reversed. Sigh.

\author{Zhenkai Qin, Baozhong Wei, Caifeng Gao, and Jianyuan Ni\textsuperscript{*}
% <-this % stops a space
\IEEEcompsocitemizethanks{\IEEEcompsocthanksitem Z. Qin, B. Wei, C. Gao are with the School of Information Technology, Guangxi Police College, Guangxi, China. \protect\\
% note need leading \protect in front of \\ to get a newline within \thanks as
% \\ is fragile and will error, could use \hfil\break instead.
\IEEEcompsocthanksitem J. Ni, is with the Department of Information Technology and Computer Science, Juniata College, Huntingdon, PA, USA. E-mail: jni100@juniata.edu.
}% <-this % stops a space
\thanks{Manuscript received December, 2024}}

% note the % following the last \IEEEmembership and also \thanks - 
% these prevent an unwanted space from occurring between the last author name
% and the end of the author line. i.e., if you had this:
% 
% \author{....lastname \thanks{...} \thanks{...} }
%                     ^------------^------------^----Do not want these spaces!
%
% a space would be appended to the last name and could cause every name on that
% line to be shifted left slightly. This is one of those "LaTeX things". For
% instance, "\textbf{A} \textbf{B}" will typeset as "A B" not "AB". To get
% "AB" then you have to do: "\textbf{A}\textbf{B}"
% \thanks is no different in this regard, so shield the last } of each \thanks
% that ends a line with a % and do not let a space in before the next \thanks.
% Spaces after \IEEEmembership other than the last one are OK (and needed) as
% you are supposed to have spaces between the names. For what it is worth,
% this is a minor point as most people would not even notice if the said evil
% space somehow managed to creep in.

% The paper headers
\markboth{Journal of \LaTeX\ Class Files, 2024}%
{Shell \MakeLowercase{\textit{et al.}}: Bare Demo of IEEEtran.cls for IEEE Journals}
\IEEEtitleabstractindextext{%
\begin{abstract}
Time series forecasting is a critical task in domains such as energy, finance, and meteorology, where accurate long-term predictions are essential. While Transformer-based models have shown promise in capturing temporal dependencies, their application to extended sequences is limited by computational inefficiencies and limited generalization. In this study, we propose KEDformer, a knowledge extraction-driven framework that integrates seasonal-trend decomposition to address these challenges. KEDformer leverages knowledge extraction methods that focus on the most informative weights within the self-attention mechanism to reduce computational overhead. Additionally, the proposed KEDformer framework decouples time series into seasonal and trend components. This decomposition enhances the model's ability to capture both short-term fluctuations and long-term patterns. Extensive experiments on five public datasets from energy, transportation, and weather domains demonstrate the effectiveness and competitiveness of KEDformer, providing an efficient solution for long-term time series forecasting.
\end{abstract}

% Note that keywords are not normally used for peerreview papers.
\begin{IEEEkeywords}
Time series forecasting, knowledge extraction, data decomposition.
\end{IEEEkeywords}}

% make the title area
\maketitle

% To allow for easy dual compilation without having to reenter the
% abstract/keywords data, the \IEEEtitleabstractindextext text will
% not be used in maketitle, but will appear (i.e., to be "transported")
% here as \IEEEdisplaynontitleabstractindextext when compsoc mode
% is not selected <OR> if conference mode is selected - because compsoc
% conference papers position the abstract like regular (non-compsoc)
% papers do!
\IEEEdisplaynontitleabstractindextext
% \IEEEdisplaynontitleabstractindextext has no effect when using
% compsoc under a non-conference mode.

% For peer review papers, you can put extra information on the cover
% page as needed:
% \ifCLASSOPTIONpeerreview
% \begin{center} \bfseries EDICS Category: 3-BBND \end{center}
% \fi
%
% For peerreview papers, this IEEEtran command inserts a page break and
% creates the second title. It will be ignored for other modes.
\IEEEpeerreviewmaketitle

\ifCLASSOPTIONcompsoc
\IEEEraisesectionheading{\section{Introduction}\label{sec:introduction}}
\else

\section{Introduction}
\label{sec:introduction}
\fi
% Computer Society journal (but not conference!) papers do something unusual
% with the very first section heading (almost always called "Introduction").
% They place it ABOVE the main text! IEEEtran.cls does not automatically do
% this for you, but you can achieve this effect with the provided
% \IEEEraisesectionheading{} command. Note the need to keep any \label that
% is to refer to the section immediately after \section in the above as
% \IEEEraisesectionheading puts \section within a raised box.

% The very first letter is a 2 line initial drop letter followed
% by the rest of the first word in caps (small caps for compsoc).
% 
% form to use if the first word consists of a single letter:
% \IEEEPARstart{A}{demo} file is ....
% 
% form to use if you need the single drop letter followed by
% normal text (unknown if ever used by the IEEE):
% \IEEEPARstart{A}{}demo file is ....
% 
% Some journals put the first two words in caps:
% \IEEEPARstart{T}{his demo} file is ....
% 
% Here we have the typical use of a "T" for an initial drop letter
% and "HIS" in caps to complete the first word.
Long-term forecasting plays a critical role in decision-making domains such as transportation logistics \cite{zhang2018long}, healthcare monitoring \cite{maray2023transfer}, utility management \cite{feng2024machine}, and energy optimization \cite{somu2020hybrid}. However, as the forecasting horizon increases, computational demands and challenges in modeling complex temporal dependencies grow substantially. Traditional time series decomposition methods, although useful, often rely on linear assumptions, making them less effective in handling complex multivariate scenarios or unpredictable, non-stationary data patterns. These limitations limit their ability to capture the interplay between components such as trends, seasonality, and irregularities \cite{gupta2024comprehensive}. 

Recent advancements have integrated deep learning approaches into the decomposition process to improve forecasting accuracy \cite{1,2}. For instance, by leveraging representation learning and nonlinear transformations, these methods aim to better capture dynamic dependencies and multi-scale interactions within time series data \cite{3,4}. More recently, transformers have excelled in various tasks, such as computer vision (CV) \cite{11,ni2024adaptive}, natural language processing (NLP) \cite{12}, and time series forecasting, due to their powerful modeling capabilities and flexibility. However, in long-term forecasting tasks, transformers still face significant challenges. For example, the computational complexity of the traditional self-attention mechanism is $O(L^2)$, where $L$ represents the sequence length. This quadratic growth leads to increasing demands on memory and computational resources, limiting the applicability of transformers in resource-constrained or real-time analysis scenarios \cite{13,14,15,17}. In addition, transformers often struggle to model long-term dependencies effectively due to noise interference, where irrelevant information weakens the attention distribution and degrades overall performance \cite{17,18}. Consequently, capturing long-term dependencies in time series data while ensuring computational efficiency over extended prediction horizons remains a significant challenge.

\begin{figure*}[t]
  \centering
  \includegraphics[width=0.8\paperwidth]{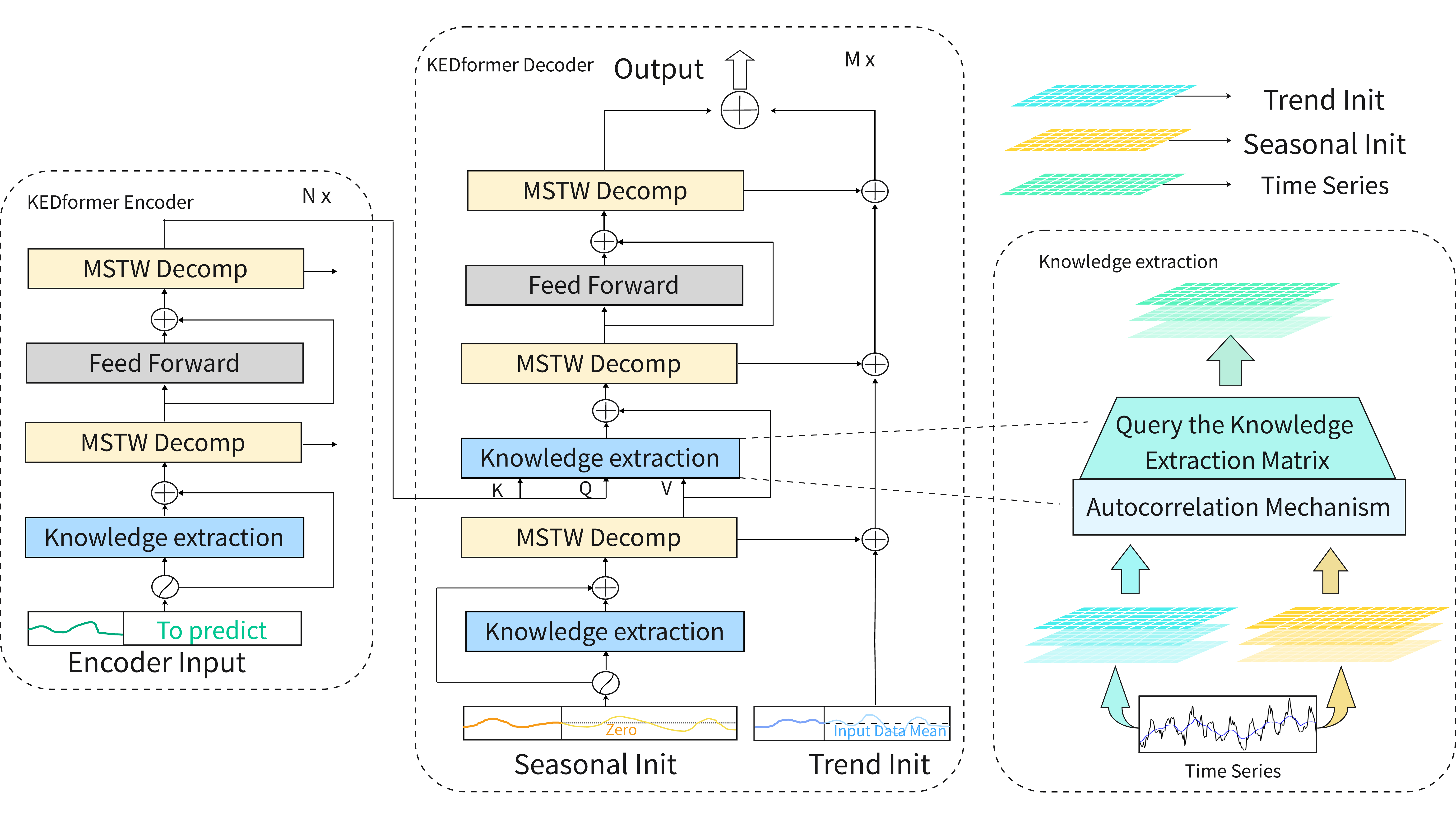}  
  \caption{Schematic overview of the proposed KEDformer method. Initially, the Knowledge Extraction Attention module (KEDA, \textcolor{blue}{blue} block) is designed to reduce model parameters by using self-correlation and sparse attention mechanisms. More specifically, the self-correlation mechanism estimates the correlation of subsequences within a specific time period, while sparse attention is employed to filter the weight matrix of these correlations. After that, the time series decomposition (MSTW, \textcolor{yellow}{yellow} block) method is used to extract seasonal and trend patterns from the input time series data.}
  \label{fig:1}
\end{figure*}

To address these challenges, we propose an end-to-end Knowledge Extraction Decomposition (KEDformer) framework for long-term time series prediction. An overview of the proposed method is shown in Figure \ref{fig:1}. First, we integrate sparse attention mechanisms with autocorrelation techniques within the model to reduce computational overhead and mitigate interference from irrelevant features. This integration reduces computational complexity from $O(L^2)$ to $O(L \log L)$, significantly lowering memory usage and enhancing the model's ability to process long sequences. Moreover, the autocorrelation mechanism decomposes the time series data into seasonal and trend components, further improving prediction accuracy. This approach captures both short-term fluctuations and long-term patterns, making the predictions more consistent with real-world temporal dynamics. As a result, the proposed KEDformer framework not only addresses the computational bottleneck of traditional Transformers in long-term forecasting, but also enhances their performance and robustness in complex sequence tasks. In summary, the contributions of this study are as follows:

\begin{itemize}
    \item We introduce a knowledge extraction mechanism that combines sparse attention and autocorrelation to reduce the computational cost of the self-attention layer. This mechanism reduces the computational overhead from quadratic to linear complexity.
    \item Furthermore, by employing seasonal-trend decomposition, KEDformer effectively captures both long-term trends and seasonal patterns, overcoming the limitations of the Transformer model in capturing long-term dependencies.
    \item Extensive experiments on five public datasets demonstrate the effectiveness and competitiveness of the proposed KEDformer, which outperforms all previous Transformer-based models across various forecasting applications.
\end{itemize}

\section{Related work}

\begin{figure*}[t]
  \centering
  \includegraphics[width=0.9\textwidth]{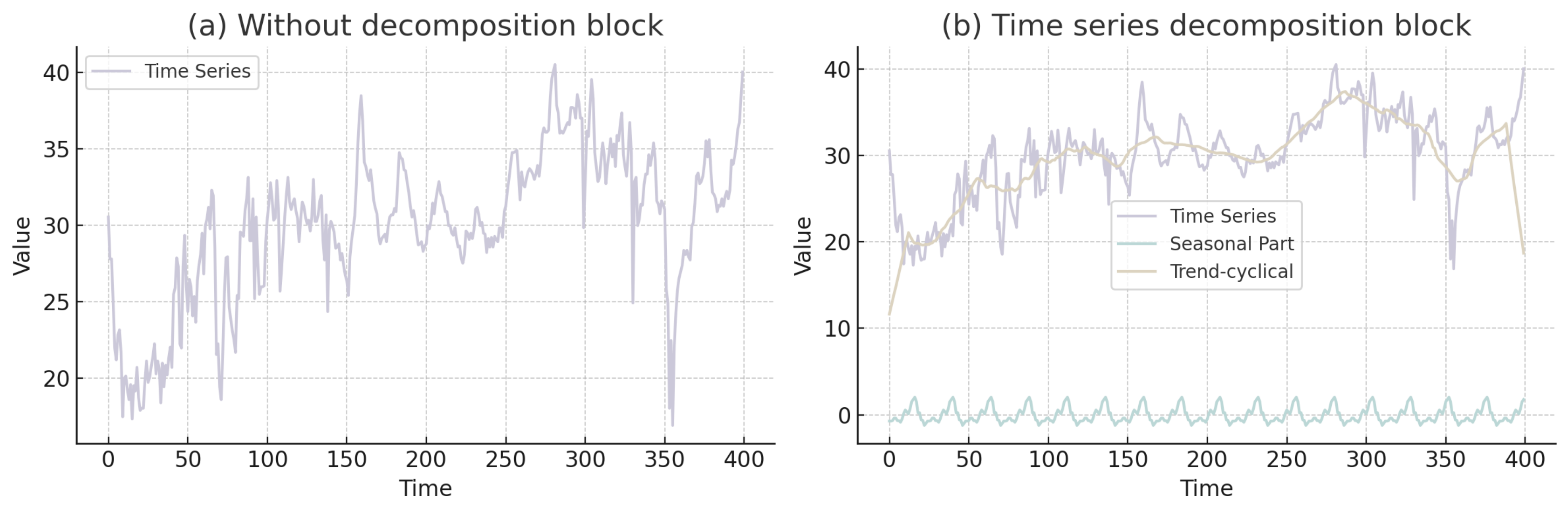}
  \caption{Visualization of time series decomposition. In the left subfigure (a), the raw time series data is shown without decomposition, displaying interwoven fluctuations and trends. In contrast, the right subfigure (b) presents the time series decomposed into three components: the original time series in \textcolor[RGB]{203,200,217}{purple}, the trend-cyclical component in \textcolor[RGB]{220,211,192}{beige}, and the seasonal component in \textcolor[RGB]{187,215,214}{teal}. Using data from the ETTm1 dataset, this decomposition reveals distinct seasonal and trend-cyclical patterns, enabling the model to better capture both periodic variations and long-term trends within the data.}
  \label{fig:2}
\end{figure*}

\subsection{Transformer-based Long-term Time Series Forecasting}

Transformer-based models have demonstrated exceptional performance in time series forecasting due to their powerful self-attention mechanism and parallel processing capabilities, excelling at capturing long-term dependencies and handling long-sequence data \cite{23,24}. However, traditional Transformer models still face several challenges in time series forecasting, such as high computational complexity and difficulty addressing noise issues in long-term dependencies \cite{12,15}. For instance, the core self-attention mechanism exhibits quadratic computational complexity with respect to sequence length, which limits its efficiency in long-sequence tasks \cite{39}.

To overcome these limitations, various advancements have been proposed in recent years. For example, Synthesizer \cite{40} investigated the importance of dot-product interactions and introduced randomly initialized, learnable attention mechanisms, demonstrating competitive performance in specific tasks. Furthermore, FNet \cite{41} replaced self-attention with Fourier transforms, showcasing its effectiveness in mixing sequence features. Another approach utilized Gaussian distributions to construct attention weights, enabling a focus on local windows and improving the performance of models in capturing local dependencies \cite{42}. Pyraformer employed a pyramid attention structure to address the complexity of handling long-range dependencies, while TFT integrated multivariate features and time-varying information to improve multi-step forecasting \cite{31}. More recently, Informer introduced the ProbSparse attention mechanism and distillation techniques, reducing computational complexity to \(O(L \log L)\) and significantly improving efficiency. LogTrans employed logarithmic sparse attention to further alleviate the computational burden of long-sequence predictions \cite{25}, while AST combined adversarial training and sparse attention to enhance robustness in complex scenarios \cite{26}. Additionally, Autoformer leveraged time series decomposition and autocorrelation mechanisms for long-term sequence forecasting \cite{13,28}, and FEDformer utilized frequency-domain enhancements to optimize performance on long sequences \cite{26,30}. Despite the progress made by these studies in optimizing computational efficiency and capturing long-term dependencies, they have shown instability in modeling complex long-term and non-periodic dependencies.

To overcome these limitations, various advancements have been proposed in recent years. For example, Synthesizer \cite{40} investigated the importance of dot-product interactions and introduced randomly initialized, learnable attention mechanisms, demonstrating competitive performance on specific tasks. Furthermore, FNet \cite{41} replaced self-attention with Fourier transforms, showcasing its effectiveness in mixing sequence features. Another approach utilized Gaussian distributions to construct attention weights, enabling a focus on local windows and improving model performance in capturing local dependencies \cite{42}. Pyraformer employed a pyramid attention structure to address the complexity of handling long-range dependencies, while TFT integrated multivariate features and time-varying information to improve multi-step forecasting \cite{31}. More recently, Informer introduced the ProbSparse attention mechanism and distillation techniques, reducing computational complexity to \(O(L \log L)\) and significantly improving efficiency. LogTrans employed logarithmic sparse attention to further alleviate the computational burden of long-sequence predictions \cite{25}, while AST combined adversarial training and sparse attention to enhance robustness in complex scenarios \cite{26}. Additionally, Autoformer leveraged time series decomposition and autocorrelation mechanisms for long-term sequence forecasting \cite{13,28}, and FEDformer utilized frequency-domain enhancements to optimize performance on long sequences \cite{26,30}. Despite the progress made by these studies in optimizing computational efficiency and capturing long-term dependencies, they have shown instability in modeling complex long-term and non-periodic dependencies.

Unlike the aforementioned studies, our proposed KEDformer integrates the selection of dominant weight distributions from sparse attention, allowing the model to better capture the intrinsic properties of time series data. Additionally, we combine sparse attention mechanisms with autocorrelation strategies to reduce computational costs while enhancing the model's stability in capturing long-term dependencies.

\subsection{Decomposition of Time Series}

Time series decomposition is a traditional approach that breaks down time series data into components such as trend, seasonality, and residuals, revealing the intrinsic patterns within the data \cite{33,34}. Among traditional methods, ARIMA \cite{kontopoulou2023review} uses differencing and parameterized modeling to decompose and forecast non-stationary time series effectively. In contrast, the Prophet model combines trend and seasonal components while accommodating external covariates \cite{33}, making it suitable for modeling complex time series patterns. Matrix decomposition-based methods, such as DeepGLO, extract global and local features through low-rank matrix decomposition, while N-BEATS employs a hierarchical structure to dissect trends and periodicity \cite{34}. However, these approaches primarily focus on the static decomposition of historical sequences and often fall short in capturing dynamic interactions for future forecasting.

More recently, deep learning models have increasingly incorporated time series decomposition to enhance predictive power. For example, Autoformer introduces an embedded decomposition module, treating trend and seasonal components as core building blocks to achieve progressive decomposition and forecasting \cite{35}. FEDformer combines Fourier and wavelet transforms to decompose time series into components of varying frequencies, capturing global characteristics and local structures while significantly reducing computational complexity and improving the accuracy of long-sequence predictions. Similarly, ETSformer \cite{36} adopts a hierarchical decomposition framework inspired by exponential smoothing, segmenting time series into level, growth, and seasonality components \cite{37}. By integrating exponential smoothing attention and frequency-domain attention mechanisms, ETSformer effectively extracts key features, demonstrating superior performance across multiple datasets \cite{38}. Inspired by these studies, our proposed KEDformer approach integrates decomposition modules dynamically with a progressive decomposition strategy. This not only significantly improves computational efficiency but also enables the simultaneous modeling of both short-term and long-term patterns.

\section{Methodology}

\subsection{Background}

The Long Sequence Time Forecasting (LSTF) problem is defined within a rolling forecasting setup, where predictions over an extended future horizon are made based on past observations within a fixed-size window \cite{39}. At each time point $t$, the input sequence $\mathcal{X}^{t} = \{x_{1}^{t}, \ldots, x_{L_{x}}^{t}\}$ consists of observations with multiple feature dimensions, and the output sequence $\mathcal{Y}^{t} = \{y_{1}^{t}, \ldots, y_{L_{y}}^{t}\}$ predicts multiple future values. The output length $L_{y}$ is intentionally set to be relatively long to capture complex dependencies over time. This setup enables the model to predict multiple attributes, making it well-suited for time series applications.

\subsection{Data Decomposition}

In this section, we will cover the following aspects of KEDformer: (1) the decomposition process designed to capture seasonal and trend components in time series data; and (2) the architecture of the KEDformer encoder and decoder.

\textbf{Time Series Decomposition} To capture complex temporal patterns in long-term predictions, we utilize a decomposition approach that separates sequences into trend, cyclical, and seasonal components. These components correspond to the long-term progression and seasonal variations inherent in the data. However, directly decomposing future sequences is impractical due to the uncertainty of future data. To address this challenge, we introduce a novel internal operation within the sequence decomposition block, referred to as the autocoupling mechanism in KEDformer, as shown in Figure \ref{fig:1}. This mechanism enables the progressive extraction of long-term stationary trends from predicted intermediate hidden states. Specifically, we adjust the moving average to smooth periodic fluctuations and emphasize the long-term trends. For the length-$L$ input sequence $\tilde{\chi} \in \mathbb{R}^{L \times d}$, the procedure is as follows:

\begin{equation}
x_t=AvgPool(Padding(X))
\end{equation}
\begin{equation}
x_s=x-x_t
\end{equation}

where \( x_{s}, x_{t} \in \mathbb{R}^{L \times d} \) represent the seasonal part and the extracted trend component, respectively. We use \( \text{AvgPool}(\cdot) \) for moving average and filling operations to maintain a constant sequence length. We summarize the above process as \( x_{s}, x_{t} = \text{MSTWDecomp}(x) \), which is a within-model block.

\textbf{Model input} In Figure \ref{fig:1}, the encoder's input consists of the past \( I \) time steps, denoted as \( X_{\text{en}} \in \mathbb{R}^{L \times d} \). In the decomposition architecture, the input to the decoder is composed of both a seasonal component, \( X_{\text{des}} \in \mathbb{R}^{(I/2+O) \times d} \), and a trend-cyclical component, \( X_{\text{det}} \in \mathbb{R}^{(I/2+O) \times d} \), both of which are subject to further refinement. Each initialization consists of two elements: (1) the decomposed component derived from the latter half of the encoder's input, \( X_{\text{en}} \), of length \( I/2 \), which provides recent information, and (2) placeholders of length \( O \), filled with scalar values. The formulation is as follows:

\begin{equation}
\mathcal{X}_{\text{ens}}, \mathcal{X}_{\text{ent}} = \operatorname{MSTWDecomp}\left(\mathcal{X}_{\text{en}\frac{I}{2}: I}\right)
\end{equation}

\begin{equation}
\mathcal{X}_{\text{des}} = \operatorname{Concat}\left(\mathcal{X}_{\text{ens}}, \mathcal{X}_{0}\right)
\end{equation}

\begin{equation}
\mathcal{X}_{\text{det}} = \operatorname{Concat}\left(\mathcal{X}_{\text{ent}}, \mathcal{X}_{\text{Mean}}\right)
\end{equation}

where \( X_{\text{ens}}, X_{\text{ent}} \in \mathbb{R}^{\frac{I}{2} \times d} \) denote the seasonal and trend-cyclical components of \( X_{\text{en}} \), respectively. The placeholders, labeled as \( X_{0}, X_{\text{Mean}} \in \mathbb{R}^{O \times d} \), are populated with zeros and the mean values of \( X_{\text{en}} \), respectively.

\textbf{Encoder} In Figure \ref{fig:1}, the encoder follows a multilayer architecture, defined as 
\[
X_{\text{en}}^l = \text{Encoder}(X_{\text{en}}^{l-1}),
\]
where \( l \in \{1, \ldots, N\} \) represents the output of the \( l \)-th encoder layer. The initial input, \( X_{\text{en}}^{0} \in \mathbb{R}^{L \times D} \), corresponds to the embedded historical time series. The Encoder function, \( \text{Encoder}(\cdot) \), is formally expressed as:

\begin{equation}
\mathcal{S}_{\text{en}}^{l, 1} = \text{ MSTWDecomp}\left(\operatorname{KEDA}\left(\mathcal{X}_{\text{en}}^{l-1}\right) + \mathcal{X}_{\text{en}}^{l-1}\right)
\end{equation}

\begin{equation}
\mathcal{S}_{\text{en}}^{l, 2} = \text{ MSWTDecomp}\left(\operatorname{FeedForward}\left(\mathcal{S}_{\text{en}}^{l, 1}\right) + \mathcal{S}_{\text{en}}^{l, 1}\right)
\end{equation}

\begin{equation}
\mathcal{X}_{\text{en}}^{l} = \mathcal{S}_{\text{en}}^{l, 2}
\end{equation}
where \( S_{\text{en}}^{l,i} \), \( i \in \{1, 2\} \) represents the seasonal component after the \( i \)-th decomposition block in the \( l \)-th layer.

\textbf{Decoder} In Figure \ref{fig:1}, the decoder has two roles: the accumulation of the trend time series part and the knowledge extraction stacking of the seasonal time series part. For example, 
\[
x_{\text{de}}^{l}, \tau_{\text{de}}^{l} = \text{Decoder}\left(x_{\text{de}}^{l-1}, \tau_{\text{de}}^{l-1}\right),
\]
where \( l \in \{1, \ldots, M\} \) represents the output of the \( l \)-th decoder layer. The decoder is formalized as:

\begin{equation}
\mathcal{S}_{d e^{\prime}}^{l, 1}\tau_{d e}^{l, 1} = \text{ MSWTDecomp}\left(\operatorname{KEDA}\left(\mathcal{X}_{d e}^{l-1}\right) + \mathcal{X}_{d e}^{l-1}\right)
\end{equation}

\begin{equation}
\mathcal{S}_{d e^{\prime}}^{l, 2}\tau_{d e}^{l, 2} = \text{ MSWTDecomp}\left(\operatorname{KEDA}\left(\mathcal{S}_{d e^{\prime}}^{l, 1}, \mathcal{X}_{e n}^{N}\right) + \mathcal{S}_{d e}^{l, 1}\right)
\end{equation}

\begin{equation}
\mathcal{S}_{d e^{\prime}}^{l, 3}\tau_{d e}^{l, 3} = \text{ MSWTDecomp}\left(\operatorname{FeedForward}\left(\mathcal{S}_{d e}^{l, 2}\right) + \mathcal{S}_{d e}^{l, 2}\right)
\end{equation}

\begin{equation}
\chi_{d e}^{l} = \mathcal{S}_{d e^{\prime}}^{l, 3}
\end{equation}

\begin{equation}
\tau_{d e}^{l} = T_{d e}^{l-1} + \mathcal{W}_{l, 1}\cdot\mathcal{T}_{d e}^{l, 1} + \mathcal{W}_{l, 2}\cdot\mathcal{T}_{d e}^{l, 2} + \mathcal{W}_{l, 3}\cdot\mathcal{T}_{d e}^{l, 3}
\end{equation}
In this context, \( S_{\text{de}}^{l,i} \) and \( T_{\text{de}}^{l,i} \), where \( i \in \{1,2,3\} \), represent the seasonal and trend components, respectively, after the \( i \)-th decomposition block within the \( l \)-th layer. The matrix \( W_{i,L} \), where \( i \in \{1,2,3\} \), serves as the projection matrix for the \( i \)-th extracted trend component, \( T_{\text{de}}^{l,i} \).
%\subsection{KEformer}

\begin{table*}[ht]
\centering
\caption{
{\bf Multivariate results.}}
\label{tab:1}
\scalebox{0.95}{\begin{tabular}{ c |c |c|c| c|c| c|c| c|c| c|c| c|c| c|c}
\hline
\multicolumn{2}{c}{\textbf{Model}} & \multicolumn{2}{|c}{\textbf{KEDformer}} & \multicolumn{2}{|c}{\textbf{Autoformer\cite{13}}} & \multicolumn{2}{|c}{\textbf{Informer\cite{14}}} & \multicolumn{2}{|c}{\textbf{Reformer\cite{16}}} & \multicolumn{2}{|c}{\textbf{LSTNet\cite{44}}} & \multicolumn{2}{|c}{\textbf{LSTM\cite{45}}} & \multicolumn{2}{|c}{\textbf{TCN\cite{46}}} \\ \hline
\multicolumn{2}{c}{\textbf{Metric}} & \multicolumn{1}{|c|}{\textbf{MSE}} & \textbf{MAE} & \multicolumn{1}{|c|}{\textbf{MSE}} & \textbf{MAE} & \multicolumn{1}{|c|}{\textbf{MSE}} & \textbf{MAE} & \multicolumn{1}{|c|}{\textbf{MSE}} & \textbf{MAE} & \multicolumn{1}{|c|}{\textbf{MSE}} & \textbf{MAE} & \multicolumn{1}{|c|}{\textbf{MSE}} & \textbf{MAE} & \multicolumn{1}{|c|}{\textbf{MSE}} & \textbf{MAE} \\ \hline
\multirow{4}{*}{\textbf{Exchange \cite{43}}} 
& \textbf{96} & \textbf{0.142} & \textbf{0.273} & \underline{0.197} & \underline{0.323} & 0.847 & 0.752 & 1.065 & 0.829 & 1.551 & 1.058 & 1.453 & 1.049 & 3.004 & 1.432 \\ \cline{2-16}
& \textbf{192} & \textbf{0.271} & \underline{0.380} & \underline{0.300} & \textbf{0.369} & 1.204 & 0.895 & 1.188 & 0.906 & 1.477 & 1.028 & 1.846 & 1.179 & 3.048 & 1.444 \\ \cline{2-16}
& \textbf{336} & \textbf{0.456} & \textbf{0.506} & \underline{0.509} & \underline{0.524} & 1.672 & 1.036 & 1.357 & 0.976 & 1.507 & 1.031 & 2.136 & 1.231 & 3.113 & 1.459 \\ \cline{2-16}
& \textbf{720} & \textbf{1.089} & \textbf{0.811} & \underline{1.447} & \underline{0.941} & 2.478 & 1.310 & 1.510 & 1.016 & 2.285 & 1.243 & 2.984 & 1.427 & 3.150 & 1.458 \\ \hline
\multirow{4}{*}{\textbf{Traffic\cite{5bdc31b817c44a1f58a0c6ab}}} 
& \textbf{96} & \textbf{0.409} & \textbf{0.379} & \underline{0.613} & \underline{0.388} & 0.719 & 0.391 & 0.732 & 0.423 & 1.107 & 0.685 & 0.843 & 0.453 & 1.438 & 0.784 \\ \cline{2-16}
& \textbf{192} & \textbf{0.607} & \textbf{0.380} & \underline{0.616} & \underline{0.382} & 0.696 & 0.379 & 0.733 & 0.420 & 1.157 & 0.706 & 0.847 & 0.453 & 1.463 & 0.794 \\ \cline{2-16}
& \textbf{336} & \textbf{0.619} & \textbf{0.323} & \underline{0.622} & \underline{0.337} & 0.777 & 0.420 & 0.742 & 0.420 & 1.216 & 0.730 & 0.853 & 0.455 & 1.479 & 0.799 \\ \cline{2-16}
& \textbf{720} & \textbf{0.656} & \textbf{0.403} & \underline{0.660} & \underline{0.408} & 0.864 & 0.472 & 0.755 & 0.423 & 1.481 & 0.805 & 1.500 & 0.804 & 1.499 & 0.804 \\ \hline
\multirow{4}{*}{\textbf{ETTm2\cite{lee2022tilde}}} 
& \textbf{96} & \textbf{0.234} & \textbf{0.316} & \underline{0.255} & \underline{0.339} & 0.365 & 0.453 & 0.658 & 0.619 & 3.142 & 1.365 & 2.041 & 1.073 & 3.041 & 1.330 \\ \cline{2-16}
& \textbf{192} & \textbf{0.278} & \textbf{0.338} & \underline{0.281} & \underline{0.340} & 0.533 & 0.563 & 1.078 & 0.827 & 3.154 & 1.369 & 2.249 & 1.112 & 3.072 & 1.339 \\ \cline{2-16}
& \textbf{336} & \textbf{0.336} & \textbf{0.369} & \underline{0.339} & \underline{0.372} & 1.363 & 0.887 & 1.549 & 0.972 & 3.160 & 1.369 & 2.568 & 1.238 & 3.105 & 1.348 \\ \cline{2-16}
& \textbf{720} & \textbf{0.417} & \textbf{0.414} & \underline{0.422} & \underline{0.419} & 3.379 & 1.388 & 2.631 & 1.242 & 3.171 & 1.368 & 2.720 & 1.287 & 3.135 & 1.354 \\ \hline
\multirow{4}{*}{\textbf{Weather\cite{627e28c55aee126c0f846893}}} 
& \textbf{96} & \textbf{0.265} & \textbf{0.333} & \underline{0.266} & \underline{0.336} & 0.332 & 0.368 & 0.689 & 0.596 & 0.594 & 0.587 & 0.560 & 0.565 & 0.615 & 0.589 \\ \cline{2-16}
& \textbf{192} & \textbf{0.305} & \textbf{0.364} & \underline{0.307} & \underline{0.367} & 0.598 & 0.544 & 0.752 & 0.638 & 0.597 & 0.587 & 0.639 & 0.608 & 0.629 & 0.600 \\ \cline{2-16}
& \textbf{336} & \textbf{0.359} & \textbf{0.399} & \underline{0.359} & \underline{0.395} & 0.702 & 0.620 & 0.639 & 0.596 & 0.597 & 0.587 & 0.455 & 0.454 & 0.639 & 0.608 \\ \cline{2-16}
& \textbf{720} & \textbf{0.414} & \textbf{0.423} & \underline{0.419} & \underline{0.428} & 0.831 & 0.731 & 1.130 & 0.792 & 0.618 & 0.599 & 0.535 & 0.520 & 0.618 & 0.599 \\ \hline
\multirow{4}{*}{\textbf{Electricity\cite{63ec4dc890e50fcafd66a4da}}} 
& \textbf{96} & \textbf{0.201} & \textbf{0.317} & \textbf{0.201} & \textbf{0.317} & \underline{0.274} & \underline{0.368} & 0.312 & 0.402 & 0.680 & 0.645 & 0.985 & 0.813 & 0.615 & 0.784 \\ \cline{2-16}
& \textbf{192} & \textbf{0.219} & \textbf{0.330} & \underline{0.222} & \underline{0.334} & 0.296 & 0.368 & 0.348 & 0.433 & 0.725 & 0.676 & 0.995 & 0.824 & 0.985 & 0.824 \\ \cline{2-16}
& \textbf{336} & \textbf{0.229} & \textbf{0.336} & \underline{0.231} & \underline{0.338} & 0.300 & 0.394 & 0.350 & 0.433 & 0.828 & 0.727 & 1.000 & 0.824 & 1.000 & 0.824 \\ \cline{2-16}
& \textbf{720} & \textbf{0.253} & \textbf{0.361} & \textbf{0.253} & \textbf{0.361} & 0.373 & 0.439 & \underline{0.340} & \underline{0.420} & 0.957 & 0.811 & 1.438 & 0.784 & 1.438 & 0.784 \\ \hline
\multicolumn{2}{c|}{\textbf{Count}} & 16 & 15 & 2 & 3 & 0 & 0 & 0 & 0 & 0 & 0 & 0 & 0 & 0 & 0 \\ \hline
\end{tabular}}
\begin{flushleft} Multivariate results with different prediction lengths $O \in \{96, 192, 336, 720\}$ for five different datasets when $I = 96$. We evaluated the performance on each dataset using two metrics: Mean Squared Error (MSE) and Mean Absolute Error (MAE), where a lower value indicates better performance. To summarize the results, we counted the number of times each model achieved the best performance. The best average results are in \textbf{bold}, while the second-best results are \underline{underlined}.
\end{flushleft}
\end{table*}

\subsection{Knowledge Extraction Process}

\subsubsection{Self-attention Mechanism}
The canonical self-attention mechanism is defined by the tuple inputs \( Q \), \( K \), and \( V \), which correspond to the query, key, and value matrices, respectively. This mechanism performs scaled dot-product attention, computed as:

\begin{equation}
A(Q, K, V) = \text{Softmax}\left(\frac{QK^T}{\sqrt{d}}\right)V
\end{equation}

where \( Q \in \mathbb{R}^{L_Q \times d} \), \( K \in \mathbb{R}^{L_K \times d} \), and \( V \in \mathbb{R}^{L_V \times d} \), with \( d \) representing the input dimension. To further analyze the self-attention mechanism, we focus on the attention distribution of the \( i \)-th query, denoted as \( q_i \), which is based on an asymmetric kernel smoother. The attention for the \( i \)-th query is formulated in probabilistic terms:

\begin{equation}
A(q_i, K, V) = \sum_j \frac{k(q_i, k_j)}{\sum_j k(q_i, k_j)} v_j = \mathbb{E}_{p(k_j|q_i)}[v_j]
\end{equation}

where \( p(k_j \mid q_i) = \frac{k(q_i, k_j)}{\sum_{j} k(q_i, k_j)} \), and \( k(q_i, k_j) \) represents the asymmetric exponential kernel \(\exp\left(\frac{q_i k_j^T}{\sqrt{d}}\right).\) This self-attention mechanism combines the values and produces outputs by computing the probability \( p(k_j \mid q_i) \). However, this process involves quadratic dot-product computations, resulting in a complexity of \( O(L_Q L_K) \), which poses a significant limitation in memory usage, particularly for models designed to enhance predictive capacity.

\subsubsection{Knowledge Selection}

From Eq. (15), the attention of the \( i \)-th query across all keys is represented as a probability distribution \( p(k_j \mid q_i) \), where the output is computed by aggregating the values \( v \) weighted by this probability. High dot-product values between query-key pairs lead to a non-uniform attention distribution, as dominant query-key pairs shift the attention probability away from a uniform distribution. If \( p(k_j \mid q_i) \) closely resembles a uniform distribution, \( q(k_j \mid q_i) = \frac{1}{L_K} \), then the self-attention essentially produces an averaged summation over the values \( v \), diminishing the significance of individual values.

To mitigate this, we introduce a knowledge extraction mechanism that evaluates the similarity between the attention probability \( p \) and a baseline distribution \( q \) using the Kullback-Leibler (KL) divergence \cite{ni2022cross, ni2022progressive}. This measure effectively reduces the influence of less significant queries. The similarity between \( p \) and \( q \) for the \( i \)-th query is computed as:

\begin{equation}
\text{KL}(q||p) = \ln \frac{1}{L_K} \sum_{j=1}^{L_K} e^{\frac{q_i k_j^T}{\sqrt{d}}} - \frac{1}{L_K} \sum_{j=1}^{L_K} \frac{q_i k_j^T}{\sqrt{d}} - \ln L_K
\end{equation}

From this, we define the distillation measure $M(q_i, K)$ for the $i$-th query as:
\begin{equation}
M(q_i, K) = \ln \left( \sum_{j=1}^{L_K} e^{\frac{q_i k_j^T}{\sqrt{d}}} \right) - \frac{1}{L_K} \sum_{j=1}^{L_K} \frac{q_i k_j^T}{\sqrt{d}}
\end{equation}

A larger \( M(q_i, K) \) value indicates that the \( i \)-th query has a more diverse attention distribution, potentially focusing on dominant dot-product pairs in the tail of the self-attention output. This approach allows the model to prioritize influential query-key pairs, thereby improving the overall effectiveness of the knowledge extraction process.

\begin{table*}[!ht]
\centering
\caption{
{\bf Univariate results.}}
\label{tab:2}
\scalebox{0.85}{\begin{tabular}{ c |c |c|c| c|c| c|c| c|c| c|c| c|c| c|c| c|c }
\hline
\multicolumn{2}{c|}{\textbf{Model}} & \multicolumn{2}{|c|}{\textbf{KEDformer}} & \multicolumn{2}{c|}{\textbf{Autoformer\cite{13}}} & \multicolumn{2}{c|}{\textbf{Informer\cite{14}}} & \multicolumn{2}{c|}{\textbf{LogTrans\cite{15}}} & \multicolumn{2}{c|}{\textbf{Reformer\cite{16}}} & \multicolumn{2}{c|}{\textbf{DeepAR\cite{48}}} & \multicolumn{2}{c|}{\textbf{Prophet\cite{taylor2018forecasting}}} & \multicolumn{2}{c}{\textbf{ARIMA\cite{49}}} \\ \hline
\multicolumn{2}{c|}{\textbf{Metric}} & \multicolumn{1}{|c|}{\textbf{MSE}} & \textbf{MAE} & \multicolumn{1}{|c|}{\textbf{MSE}} & \textbf{MAE} & \multicolumn{1}{|c|}{\textbf{MSE}} & \textbf{MAE} &\multicolumn{1}{|c|}{\textbf{MSE}} & \textbf{MAE} & \multicolumn{1}{|c|}{\textbf{MSE}} & \textbf{MAE} & \multicolumn{1}{|c|}{\textbf{MSE}} & \textbf{MAE} & \multicolumn{1}{|c|}{\textbf{MSE}} & \textbf{MAE} & \multicolumn{1}{|c|}{\textbf{MSE}} & \textbf{MAE} \\ \hline
\multirow{4}{*}{\textbf{ETTm2\cite{lee2022tilde}}} 
& \textbf{96} & \textbf{0.064} & \textbf{0.187} & \underline{0.065} & \underline{0.189} & 0.088 & 0.225 & 0.082 & 0.217 & 0.131 & 0.288 & 0.099 & 0.253 & 0.287 & 0.456 & 0.211 & 0.362 \\ \cline{2-18}
& \textbf{192} & \textbf{0.114} & \textbf{0.251} & \underline{0.118} & \underline{0.256} & 0.132 & 0.283 & 0.133 & 0.284 & 0.186 & 0.354 & 0.154 & 0.304 & 0.312 & 0.483 & 0.261 & 0.406 \\ \cline{2-18}
& \textbf{336} & \textbf{0.147} & \textbf{0.297} & \underline{0.147} & \underline{0.305} & 0.180 & 0.336 & 0.201 & 0.361 & 0.220 & 0.381 & 0.277 & 0.428 & 0.428 & 0.593 & 0.317 & 0.448 \\ \cline{2-18}
& \textbf{720} & \textbf{0.181} & \textbf{0.333} & \underline{0.182} & \underline{0.335} & 0.300 & 0.435 & 0.269 & 0.407 & 0.267 & 0.430 & 0.332 & 0.468 & 0.534 & 0.593 & 0.366 & 0.487 \\ \hline
\multirow{4}{*}{\textbf{Exchange\cite{43}}} 
& \textbf{96} & \underline{0.161} & \underline{0.309} & 0.241 & 0.387 & 0.591 & 0.615 & 0.279 & 0.441 & 1.327 & 0.944 & 0.417 & 0.515 & 0.828 & 0.762 & \textbf{0.112} & \textbf{0.245} \\ \cline{2-18}
& \textbf{192} & \textbf{0.203} & \textbf{0.356} & \underline{0.273} & \underline{0.403} & 1.183 & 0.912 & 0.315 & 0.498 & 1.258 & 0.924 & 0.813 & 0.735 & 0.909 & 0.974 & 0.304 & 0.404 \\ \cline{2-18}
& \textbf{336} & \textbf{0.489} & \textbf{0.497} & \underline{0.508} & \underline{0.539} & 1.367 & 0.984 & 2.438 & 1.048 & 1.262 & 1.296 & 1.331 & 0.962 & 1.304 & 0.988 & 0.736 & 0.598 \\ \cline{2-18}
& \textbf{720} & \textbf{0.896} & \textbf{0.724} & \underline{0.991} & \underline{0.768} & 1.872 & 1.072 & 2.010 & 1.181 & 1.280 & 0.953 & 1.894 & 1.181 & 3.238 & 1.566 & 1.871 & 0.935 \\ \hline
\multicolumn{2}{c|}{\textbf{Count}} & 7 & 7 & & 0 & 0 &  0 & 
0 & 0 & 0 & 0 & 0 & 0 & 0 & 0 & 1 & 1  \\ \hline
\end{tabular}}
\begin{flushleft} Univariate results with different prediction lengths \( O \in \{96, 192, 336, 720\} \) for two different datasets when \( I = 96 \). We calculate the Mean Squared Error and the Mean Absolute Error for each dataset. A lower value indicates better performance. The best average results are in \textbf{bold}, while the second-best results are in \underline{underlined}.The percentage reduction in error signifies the extent of performance enhancement when transitioning from one model to another.
\end{flushleft}
\end{table*}

\begin{table*}[h]
    \centering
    \caption{
    {\bf Ablation study of KEDformer and Its variants for multivariate long-term time series forecasting.}}
    \label{tab:3}
    \begin{tabular}{c |c |c|c| c|c| c|c| c|c| c|c}
        \hline
        \multicolumn{2}{c|}{\textbf{Model}} & \multicolumn{2}{c|}{\textbf{KEDformer}} & \multicolumn{2}{c|}{\textbf{KEDformerV1}} & \multicolumn{2}{c|}{\textbf{KEDformerV2}} & \multicolumn{2}{c|}{\textbf{Informer\cite{14}}} & \multicolumn{2}{c}{\textbf{Reformer\cite{16}}} \\ \hline
        \multicolumn{2}{c|}{\textbf{Self-att}}  & \multicolumn{2}{c|}{KEDatt} & \multicolumn{2}{c|}{KEDatt} & \multicolumn{2}{c|}{KEDatt-f} & \multicolumn{2}{c|}{ProbAtt} & \multicolumn{2}{c}{Reatt} \\ \hline
        \multicolumn{2}{c|}{\textbf{Cross-att}} & \multicolumn{2}{c|}{KEDatt} & \multicolumn{2}{c|}{KEDatt-f} & \multicolumn{2}{c|}{KEDatt-f} & \multicolumn{2}{c|}{ProbAtt} & \multicolumn{2}{c}{Reatt} \\  \hline
        \multicolumn{2}{c|}{\textbf{Metric}}  & \textbf{mse} & \textbf{mae} & \textbf{mse} & \textbf{mae} & \textbf{mse} & \textbf{mae} & \textbf{mse} & \textbf{mae} & \textbf{mse} & \textbf{mae} \\ 
        \hline
        \multirow{4}{*}{\textbf{Exchange\cite{43}}} 
        & \textbf{96} & \textbf{0.142} & \textbf{0.273} & 0.175 & 0.318 & \underline{0.170} & \underline{0.315} & 0.847 & 0.752 & 1.065 & 0.829 \\ \cline{2-12}
        & \textbf{192} & \textbf{0.271} & \textbf{0.380} & \underline{0.281} & \underline{0.387} & 0.302 & 0.407 & 1.204 & 0.895 & 1.188 & 0.906 \\ \cline{2-12}
        & \textbf{336} & \textbf{0.456} & \textbf{0.506} & \underline{0.472} & \underline{0.514} & 0.502 & 0.535 & 1.672 & 1.036 & 1.357 & 0.976 \\ \cline{2-12}
        & \textbf{720} & \underline{1.089} & \underline{0.811} & \textbf{1.094} & \textbf{0.798} & 1.097 & 0.821 & 2.478 & 1.310 & 1.510 & 1.016 \\ 
        \hline
        \multirow{4}{*}{\textbf{Weather\cite{627e28c55aee126c0f846893}}} 
        & \textbf{96} & \textbf{0.265} & \textbf{0.333} & \underline{0.277} & \underline{0.360} & 0.354 & 0.382 & 0.384 & 0.458 & 0.689 & 0.596 \\ \cline{2-12}
        & \textbf{192} & \textbf{0.305} & \textbf{0.364} & 0.418 & 0.467 & \underline{0.359} & \underline{0.417} & 0.544 & 0.652 & 0.752 & 0.638 \\ \cline{2-12}
        & \textbf{336} & \textbf{0.359} & \textbf{0.399} & \underline{0.482} & \underline{0.505} & 0.518 & 0.523 & 0.794 & 0.794 & 0.639 & 0.596 \\ \cline{2-12}
        & \textbf{720} & \textbf{0.414} & \textbf{0.423} & \underline{0.554} & \underline{0.544} & 0.645 & 0.604 & 0.741 & 0.869 & 1.130 & 0.792 \\ \hline
        \multicolumn{2}{c|}{\textbf{Count}} & 7 & 7 & 1 & 1 & 0 & 0 & 0 & 0 & 0 & 0 \\ \hline
    \end{tabular}
    \begin{flushleft} Multivariate long-term time series forecasting results on the Exchange and Weather datasets, with an input length $I=96$ and prediction lengths $O \in \{96, 192, 336, 720\}$, are presented. Two variants of KEDformer are compared with baselines. We evaluated performance on each dataset using two metrics: Mean Squared Error (MSE) and Mean Absolute Error (MAE), where a lower value indicates better performance. To summarize the results, we counted the number of times each model achieved the best performance. The best average results are in \textbf{bold}, while the second-best results are \underline{underlined}.
    \end{flushleft}
\end{table*}

\subsubsection{Decoupled Knowledge Extraction}

\textbf{Period-based dependencies} The period-based dependencies are quantified using the autocorrelation function, which measures the similarity between different time points in a time series, revealing its underlying periodic characteristics. For a discrete time series \(\{X_t\}\), the autocorrelation function is defined as:

\begin{equation}
R_{XX}(\tau)=\lim_{L\rightarrow\infty}\frac{1}{L}\sum_{t=1}^{L} X_t X_{t-\tau},
\end{equation}

Where $\tau$ represents the time lag, $L$ is the total length of the series, and $\{X_t\}$ and $\{X_{t-\tau}\}$ are the values at the current and lagged time points, respectively. The autocorrelation function computes the cumulative similarity over lagged time intervals, reflecting the degree of self-similarity within the series for various time delays. Peaks in the autocorrelation values indicate potential periodicity and help identify the likely period lengths.

By identifying the peaks of the autocorrelation function, the most probable period lengths \((\tau_1, \tau_2, \ldots, \tau_k)\) can be determined. These period lengths not only capture the dominant periodic patterns in the series but also serve as weighted features, enhancing interpretability and predictive capabilities.

\begin{figure*}[t]
  \centering
  \includegraphics[width=1.0\textwidth]{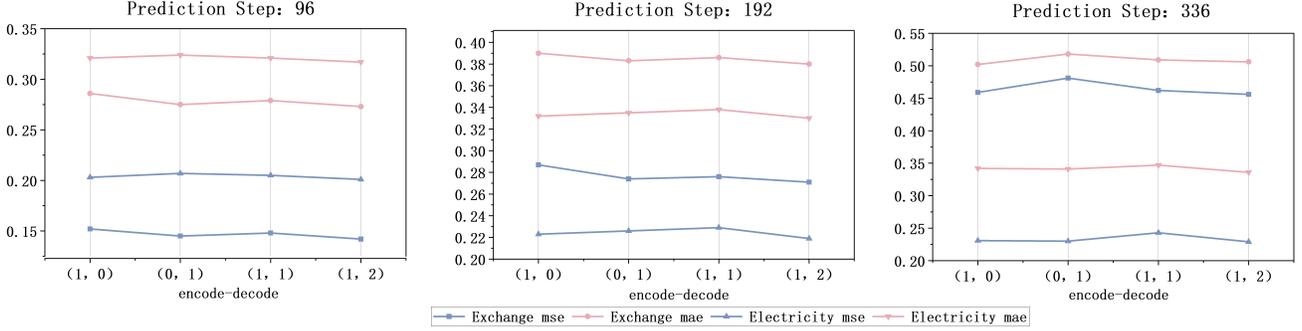}
  \caption{The results of the time series decomposition experiment are presented. In a comparative experiment that controls the number of KEDformer mechanisms during the encoding and decoding processes, we set the input length $I=96$ and the prediction lengths $O \in \{96, 192, 336\}$. The evaluation metrics used are Mean Squared Error (MSE) and Mean Absolute Error (MAE), with lower values indicating better model performance.}
  \label{fig3}
\end{figure*}

\textbf{Time-delay Aggregation} 
The time-delay aggregation method for knowledge acquisition focuses on estimating the correlation of sub-sequences within a specific period. Therefore, we propose an innovative time-delay aggregation module that can perform hierarchical convolution operations on sub-sequences based on the selected time delays $\tau_{1}, \ldots, \tau_{k}$, thereby narrowing down the key knowledge weight matrix. This process captures sub-sequences from the same location and similar positions within the period, extracting the potential key-weight aggregation matrix. Finally, we apply the Softmax function to normalize the weights, enhancing the accuracy of sub-sequence aggregation.

For a time series \( x \) of length \( L \), after projection and filtering of the weight matrix, we obtain the query \( \hat{Q} \), key \( K \), and value \( v \). The knowledge extraction attention mechanism is then as follows:

\begin{equation}
\tau_1,\cdots,\tau_k = \arg\operatorname{Topk}_{\tau\in\{1,\cdots,L\}}\left(\mathcal{R}_{\mathcal{Q},\mathcal{K}}(\tau)\right)
\end{equation}

\begin{equation}
\mathcal{R}_{\widehat{Q},\mathcal{K}}(\tau) = \operatorname{Top}_u(M(Q,K)) \cdot \mathcal{R}_{Q,\mathcal{K}}(\tau)
\end{equation}

\begin{equation}\small
\mathcal{R}_{\widehat{Q},\mathcal{K}}(\tau_{1}),\cdots,\mathcal{R}_{\widehat{Q},\mathcal{K}}(\tau_{k}) = \operatorname{SoftMax}\left(\mathcal{R}_{\widehat{Q},\mathcal{K}}(\tau_{1}),\cdots,\mathcal{R}_{\widehat{Q},\mathcal{K}}(\tau_{k})\right)
\end{equation}

\begin{equation}
\text{KEDattention}(\widehat{Q},\mathcal{K},\mathcal{V}) = \sum_{i=1}^{k} \operatorname{Roll}(\mathcal{V},\tau_{i}) \mathcal{R}_{\widehat{Q},\mathcal{K}}(\tau_{i})
\end{equation}

Where \(\arg\operatorname{Topk}(\cdot)\) is used to obtain the top \(k\) parameters of self-attention, and let \(k = \lceil c \times \log L \rceil\), where \(c\) is a hyperparameter. \(\mathcal{R}_{Q, K}\) represents the self-attention matrix between sequences \(Q\) and \(K\). \(Top_{u}\) selects the most important \(u\) queries in the weight matrix. \(\mathcal{R}_{\hat{Q},\mathcal{K}}\) represents the self-attention matrix after filtering between sequences \(Q\) and \(K\). \(\operatorname{Roll}(X, \tau)\) denotes the operation of temporally shifting \(X\) by \(\tau\), where the elements shifted out from the front are reintroduced at the end. For the encoder-decoder self-attention, \(K\) and \(V\) come from the encoder \(X_{\text{en}}\) and are adjusted to length \(O\), with \(Q\) originating from the previous block of the decoder.

\begin{figure*}[t]
\centering
  \includegraphics[width=1.0\textwidth]{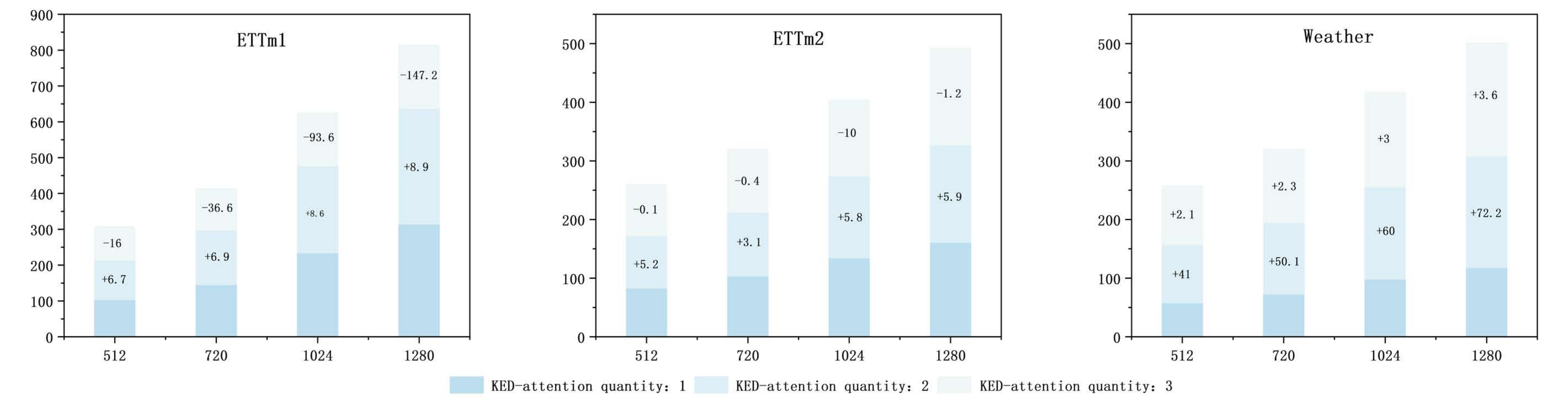}
  \caption{The number of KEDformer mechanisms and their impact on the computational efficiency of the model are evaluated by controlling the number of mechanisms. The input length is set to $I=96$, and the prediction steps are $O=\{512, 720, 1024, 1280\}$. The time required for each epoch is used as an indicator of the model's computational speed.}
  \label{fig4}
\end{figure*}

\section{Experiment}

\subsection{Datasets}

Five public datasets across multiple tasks were used to evaluate the effectiveness of the proposed KEDformer model as follows: (1) ETT \cite{lee2022tilde}: This dataset comprises four sub-datasets—ETTh1, ETTh2, ETTm1, and ETTm2. The data in ETTh1 and ETTh2 were sampled every hour, while the data in ETTm1 and ETTm2 were sampled every 15 minutes. These datasets include load and oil temperature measurements collected from power transformers between July 2016 and July 2018. (2) Electricity \cite{63ec4dc890e50fcafd66a4da}: This dataset contains hourly electricity consumption data from 321 customers, spanning from 2012 to 2014. (3) Exchange Rates \cite{43}: This dataset records daily exchange rates across eight different countries from 1990 to 2016. (4) Traffic \cite{5bdc31b817c44a1f58a0c6ab}: This dataset consists of hourly traffic data from the California Department of Transportation, capturing road occupancy through various sensors on the Bay Area Highway. (5) Weather \cite{627e28c55aee126c0f846893}: This dataset includes meteorological data recorded every 10 minutes throughout the year 2020, with 21 indicators such as temperature and humidity. In accordance with standard protocols, all datasets were chronologically split into training, validation, and test sets. The ETT dataset was partitioned using a 6:2:2 ratio \cite{lee2022tilde}, while the other datasets followed a 7:1:2 split \cite{63ec4dc890e50fcafd66a4da,43,5bdc31b817c44a1f58a0c6ab,627e28c55aee126c0f846893}.

\subsection{Implementation Details}

For the Transformer model, due to its approach to handling time series data, residual connections are embedded within the decomposition blocks \cite{28}. These blocks include processes such as moving averages to smooth out periodic fluctuations and highlight long-term trends in the data. By incorporating residual connections in this manner, the model is better equipped to learn and leverage the complex patterns within time series, thereby improving performance in long-term time series forecasting. Our method is trained using L2 loss with the ADAM \cite{5550415745ce0a409eb3a739} optimizer. The entire training process is initialized with a fixed random seed. The initial learning rate is set to \(10^{-4}\), with a batch size of 32. The attention factor is set to 3, and the weight decay is set to 0.1. Training is stopped early after 10 epochs. All experiments are repeated three times and implemented in PyTorch \cite{Paszke2019PyTorch}, running on a single NVIDIA Tesla V100 32GB GPU \cite{5aed14e217c44a4438159857}.

We evaluated nine baseline methods for comparison. For the multivariate setting, we selected three Transformer-based models: Autoformer \cite{28}, which introduces decomposition blocks for trend-seasonality extraction and employs auto-correlation mechanisms to effectively capture long-range dependencies; Informer \cite{14}, which enhances performance in processing long-sequence data through its probabilistic sparse self-attention mechanism and self-attention distillation; and Reformer \cite{16}, which optimizes computational efficiency and memory usage using Locality-Sensitive Hashing (LSH) and Reversible Layers. Additionally, we included two RNN-based models: LSTNet \cite{44}, which leverages adaptive feature selection and multi-scale forecasting for improved long-term time series prediction, and LSTM \cite{45}, which captures long-term dependencies using gated mechanisms. For CNN-based models, we selected TCN \cite{46}, designed to capture local patterns and long-range dependencies in time series data through causal and dilated convolutions.

In the univariate setting, we incorporated several competitive baselines: LogTrans \cite{47}, which improves the efficiency and accuracy of Transformers in time series forecasting with convolutional self-attention and sparse biases; DeepAR \cite{48}, which enhances forecasting accuracy by learning complex patterns, including seasonality and trends, through deep learning techniques; Prophet \cite{taylor2018forecasting}, a model that combines statistical methods and machine learning to effectively handle strong seasonal patterns and multiple seasonal cycles; and ARIMA \cite{49}, which integrates autoregressive (AR), integrated (I), and moving average (MA) components to effectively capture and forecast trends and seasonal patterns in time series data.

\subsubsection{Performance Comparison}

\textbf{Multivariate results}: In multivariate settings, KEDformer consistently demonstrated superior performance across all benchmarks, as shown in Table \ref{tab:1}. Notably, under the input-96-predict-336 configuration, KEDformer achieved significant improvements across five real-world datasets, while its predictive performance on the weather dataset remained unchanged. The mean squared error (MSE) was notably reduced by 0.8\% (0.339 $\rightarrow$ 0.336) in the ETT dataset, by 0.8\% (0.231 $\rightarrow$ 0.229) in the Electricity dataset, by 10.4\% (0.509 $\rightarrow$ 0.456) in the Exchange dataset, and by 0.4\% (0.622 $\rightarrow$ 0.619) in the Traffic dataset. On average, KEDformer reduced MSE by 2.48\% across these datasets. In particular, KEDformer demonstrated substantial improvements on the Exchange dataset, which is characterized by a lack of apparent periodicity. Moreover, the model's performance remained stable as the prediction length increased, indicating its robustness in long-term forecasting. This robustness is especially beneficial for practical applications, such as early weather warning systems and long-term energy consumption planning.

\noindent\textbf{Univariate results}: 
We present the univariate results for two representative datasets, as shown in Table \ref{tab:2}. Compared to various baseline models, KEDformer has largely achieved state-of-the-art performance in long-term prediction tasks. Specifically, in the input-96-predict-336 configuration, our model reduces the Mean Absolute Error (MAE) on the ETTm2 dataset by 2.6\% ($0.305 \rightarrow 0.297$). In the power dataset, which exhibits significant periodicity, KEDformer demonstrates its effectiveness. For the Exchange dataset, which lacks significant periodicity, KEDformer outperformed other baselines by 7.7\% ($0.539 \rightarrow 0.497$), thereby demonstrating excellent long-term predictive power. Additionally, ARIMA achieved the best performance in the input-96-predict-96 configuration on the Exchange dataset; however, its performance declined in long-term forecasting settings. This decline can be attributed to ARIMA's strong ability to capture global features in time series data during processing. However, it is constrained by the complex temporal patterns in real-world time series, which pose challenges for long-term predictions.

\begin{figure*}[t]
  \centering
  \includegraphics[width=1.0\textwidth]{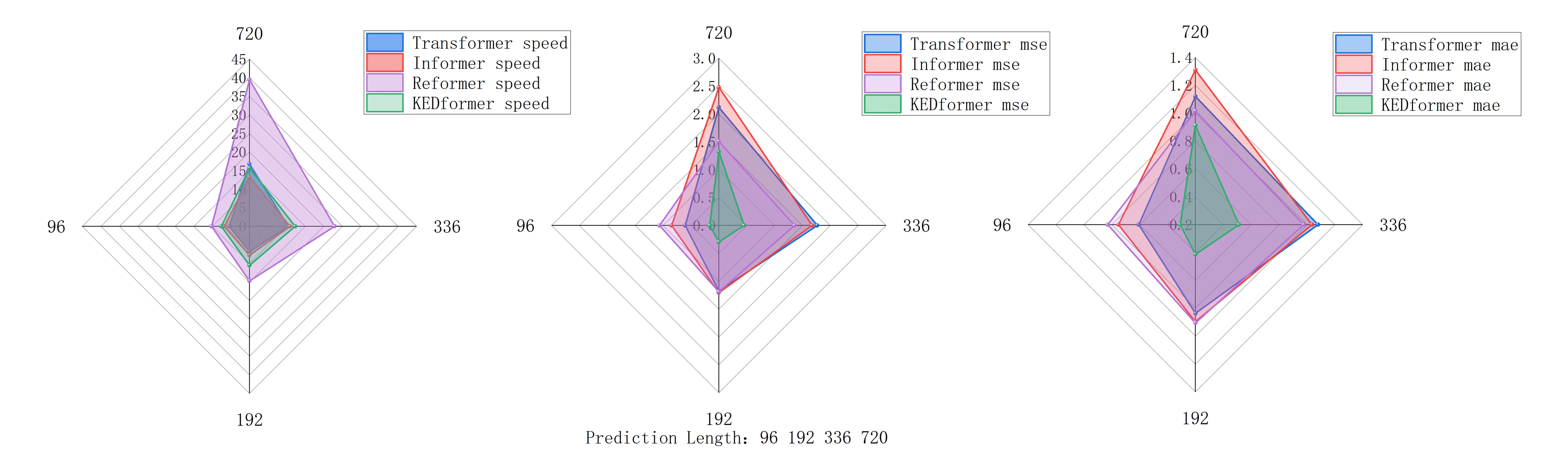}
  \caption{In the experiment for model computational efficiency and performance analysis, four different models are used to perform long-term time series forecasting tasks on the Exchange dataset. The input length is set to $I=96$, and the prediction lengths are $O \in \{96, 192, 336, 720\}$. The metric for evaluating computational efficiency is the time (in seconds) taken by each model to compute one epoch, while the performance metrics are the Mean Squared Error (MSE) and Mean Absolute Error (MAE).}
  \label{fig5}
\end{figure*}

\subsection{Ablation research}
This study evaluates the impact of the Knowledge Extraction Attention (KEDA) module on model performance through ablation experiments, testing three KEDformer variants: KEDformer, which completely replaces both the self-attention and cross-attention mechanisms with KEDA; KEDformer V1, which replaces only the self-attention mechanism with KEDA while retaining traditional attention for cross-attention; and KEDformer V2, which uses traditional attention for both mechanisms. Experiments were conducted on the Exchange and Weather datasets, as shown in Table \ref{tab:3}, where KEDformer achieved performance improvements in 14 out of 16 test cases, whereas KEDformer V1 showed enhancements in only 2 cases. Notably, KEDformer with the KEDA module demonstrated consistent improvements across all cases, confirming the effectiveness of KEDA in replacing attention mechanisms and significantly enhancing model performance.

\subsubsection{Time series decomposition effects on the mode}
The integration of time series decomposition into the KEDformer model substantially improves predictive accuracy by isolating seasonal patterns from trend components. This decomposition allows the model to focus on short-term fluctuations while preserving an understanding of long-term trends, which is crucial for accurate forecasting. As demonstrated in Figure \ref{fig:2}, this enhancement can be attributed to several factors. First, by explicitly modeling seasonal variations, the model can adapt more effectively to recurring patterns, thus improving its ability to project future values based on historical data. Second, decomposition helps identify significant features within the data, enabling the model to prioritize relevant information during the forecasting process.

\subsubsection{Effect of KEDformer number on Encoder and Decoder}
In this study, we conducted comparative experiments using the Exchange dataset, varying the number of KEDformer mechanisms. The results, illustrated in Figure \ref{fig3}, demonstrate that the model achieves superior performance when the number of KEDformer mechanisms in the decoding phase exceeds that in the encoding phase. This improvement can be attributed to the model's enhanced ability to focus on the most informative features during the decoding process, effectively capturing dependencies between the predicted outputs and historical inputs. Conversely, performance declines when the number of KEDformer mechanisms in the decoding phase equals that in the encoding phase.

\subsubsection{Effect of KEDformer on computational efficiency}
We conducted experiments to evaluate the impact of increasing the number of KEDformer mechanisms on the computational efficiency of the model, as shown in Figure \ref{fig4}. The results demonstrate that the model achieves improved efficiency as the number of KEDformer mechanisms increases across various datasets (ETTm1, ETTm2, and Weather). Notably, the time required for each epoch decreases significantly with the increase in the number of KEDformer mechanisms, with the most pronounced improvement observed in the ETTm1 dataset, where computation time drops from 794.0 seconds to 467.2 seconds. This enhancement can be attributed to the model's improved ability to capture temporal dependencies and optimize resource utilization, enabling parallel processing and more effective distribution of the computational load.

\subsubsection{Efficiency analysis and performance analysis}
In this study, we conducted an efficiency and performance analysis of models utilizing different self-attention mechanisms, as illustrated in Figure \ref{fig5}. On the Exchange dataset, the KEDformer model ranked third in terms of running time but achieved the highest prediction accuracy, thanks to its optimized knowledge extraction mechanism and seasonal trend decomposition approach. These features enhance the model's ability to capture key patterns in the time series. However, the model's performance may degrade when handling time series data without clear periodicity, as the seasonal trend decomposition may fail to effectively extract relevant information. Additionally, an inappropriate configuration of the number of KEDformer mechanisms can reduce efficiency and negatively impact the final prediction results.

\section{Conclusion}

In this study, we introduce KEDformer, a novel framework designed to address the computational inefficiencies inherent in the self-attention mechanism for long-term time series forecasting. By leveraging sparse attention, KEDformer reduces computational complexity from quadratic to linear time, significantly improving processing speed. The framework integrates seasonal-trend decomposition and autocorrelation mechanisms, which greatly enhance the model's ability to capture both short-term fluctuations and long-term patterns in time series data. This dual approach minimizes information loss during prediction, enabling the model to more effectively align with real-world time dynamics. This alignment is particularly critical for accurate forecasting in domains such as energy, finance, and meteorology. Experimental results across several benchmark datasets demonstrate that KEDformer consistently outperforms existing Transformer-based models, highlighting its robustness and adaptability. These findings underscore the potential of KEDformer as a valuable tool for long-term time series forecasting.

Despite these advancements, we acknowledge several limitations. First, when applied to non-periodic datasets, the effectiveness of KEDformer may decrease, as seasonal-trend decomposition and autocorrelation mechanisms may not provide significant benefits in such cases. Moreover, optimizing the weight balancing within the self-attention mechanism remains a challenging issue that requires further investigation. Future research will focus on enhancing KEDformer's adaptability to a wider range of datasets, including those with irregular patterns. We also aim to refine the knowledge extraction process to extend the framework’s applicability to other sequence-based tasks, broadening its practical utility beyond time series forecasting. In conclusion, KEDformer represents a significant advancement in addressing the challenges of long-term time series forecasting, offering a solution that balances both efficiency and accuracy.

\vfill

% Can be used to pull up biographies so that the bottom of the last one
% is flush with the other column.
%\enlargethispage{-5in}

\vfill\pagebreak
\bibliographystyle{IEEEtran}
\bibliography{sample-base}
\end{document}